\pdfoutput=1

\documentclass[11pt]{article}

\usepackage[final]{acl}

\usepackage{amsmath}
\usepackage{times}
\usepackage{latexsym}
\usepackage{booktabs}
\usepackage{float}
\usepackage{multirow}
\usepackage{xspace}
\usepackage{wasysym}
\usepackage{amssymb}
\usepackage{pifont}
\usepackage{xcolor}         
\usepackage{colortbl}
\usepackage{color}
\usepackage{makecell}
\usepackage{geometry}
\usepackage{mathtools}
\usepackage[most, breakable, many]{tcolorbox}
\usepackage{ulem}

\definecolor{Gray}{gray}{0.90}

\definecolor{mypurple}{HTML}{25004D}
\definecolor{darkgreen}{HTML}{2D8659}
\definecolor{darkred}{HTML}{990000}

\usepackage[T1]{fontenc}

\usepackage[utf8]{inputenc}

\usepackage{microtype}

\usepackage{inconsolata}

\usepackage{graphicx}

\usepackage{subfigure}
\usepackage{enumitem}
\newenvironment{itemize*}%
 {\leftmargini=20pt\begin{itemize}%
  \setlength{\itemsep}{3pt}%
  \setlength{\parskip}{0pt}%
  }%
 {\end{itemize}}
\newenvironment{enumerate*}%
 {\begin{enumerate}%
  \setlength{\itemsep}{0pt}%
  \setlength{\parskip}{0pt}}%
 {\end{enumerate}}

\title{
Can a Single Model Master Both Multi-turn Conversations and Tool Use?\\
CoALM: A Unified Conversational Agentic Language Model 
}

\author{
Emre Can Acikgoz$^{1}$, Jeremiah Greer$^{2}$, Akul Datta$^{1}$, Ze Yang$^{1}$, William Zeng$^{2}$,\\
\textbf{Oussama Elachqar$^{2}$, Emmanouil Koukoumidis$^{2}$, Dilek Hakkani-Tür$^{1}$, Gokhan Tur$^{1}$}\\
$^{1}$University of Illinois Urbana-Champaign, $^{2}$Oumi\\
\texttt{\{acikgoz2, akuld2, zey2, dilek, gokhan\}@illinois.edu}\\
\texttt{\{jeremy, william, oussama, manos\}@oumi.ai}\\
}

\begin{document}
\maketitle
\begin{abstract}
Large Language Models (LLMs) with API-calling capabilities enabled building effective Language Agents (LA), while also revolutionizing the conventional task-oriented dialogue (TOD) paradigm.
However, current approaches face a critical dilemma: TOD systems are often trained on a limited set of target APIs, requiring new data to maintain their quality when interfacing with new services, while LAs are not trained to maintain user intent over multi-turn conversations. 
Because both robust multi-turn management and advanced function calling are crucial for effective conversational agents, we evaluate these skills on three popular benchmarks: MultiWOZ 2.4 (TOD), BFCL V3 (LA), and API-Bank (LA)—and our analyses reveal that specialized approaches excel in one domain but underperform in the other.
To bridge this chasm, we introduce \textbf{CoALM} (\uline{Co}nversational \uline{A}gentic \uline{L}anguage \uline{M}odel), a unified approach that integrates both conversational and agentic capabilities.
We created \textbf{CoALM-IT}, a carefully constructed multi-task dataset that interleave multi-turn ReAct reasoning with complex API usage. 
Using CoALM-IT, we train three models \textbf{CoALM 8B}, \textbf{CoALM 70B}, and \textbf{CoALM 405B}, which outperform top domain-specific models, including GPT-4o, across all three benchmarks. This demonstrates the feasibility of a single model approach for both TOD and LA, setting a new standard for conversational agents\footnote{\url{https://emrecanacikgoz.github.io/CoALM/}}.

\end{abstract}


\section{Introduction}



The concept of intelligent agents has been the cornerstone of artificial intelligence research for a long time \cite{Minsky1986-MINTSO}, developing in parallel with the field of human-to-machine conversation \cite{young02_icslp}. The advent of LLMs \cite{Achiam2023GPT4TR, Dubey2024TheL3-llama3} has revolutionized both fields and enabled powerful Language Agents (LA) \cite{schick2024toolformer} while transforming modular dialogue systems into end-to-end solutions \cite{hudecek-dusek-2023-large}. Despite sharing LLM foundations, they are typically focused and analyzed separately from each other; dialogue models focused on tasks such as multi-turn interactions, delivering relevant information to users, and dialogue management with state-tracking, on the other hand LAs concentrated exclusively on tool calling skills.

\begin{figure}[t!]
\includegraphics[width=\linewidth]{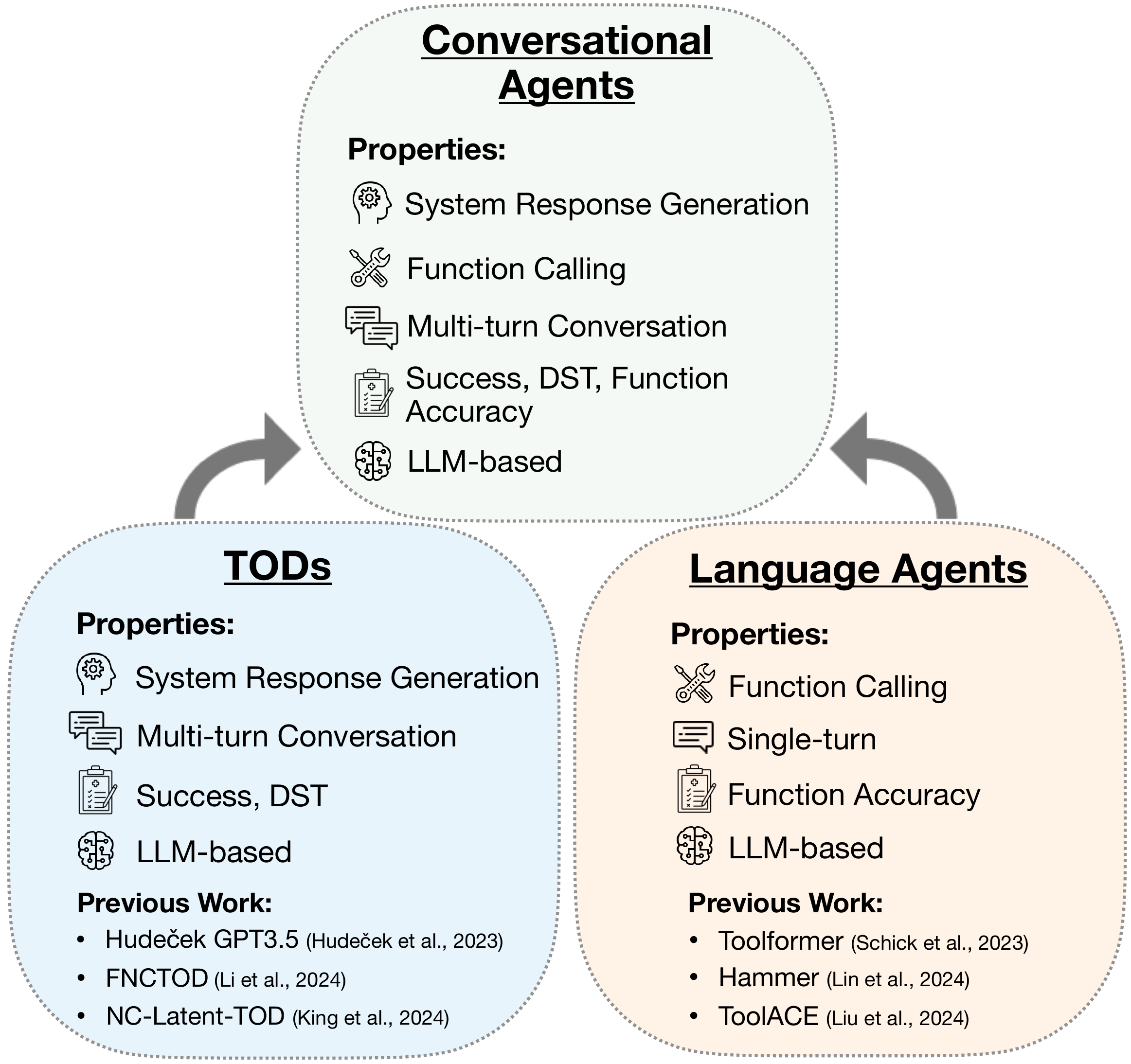}
\caption{\textbf{Unifying Capabilities of TOD Systems and LAs.} TOD systems excel in multi-turn conversations and task completion but lack advanced API capabilities, while LA handle APIs well but struggle with coherent multi-turn dialogue.}
\vspace{-5mm}
\label{fig:conv-agent}
\end{figure}

\begin{figure*}[t!]
\includegraphics[width=\linewidth]{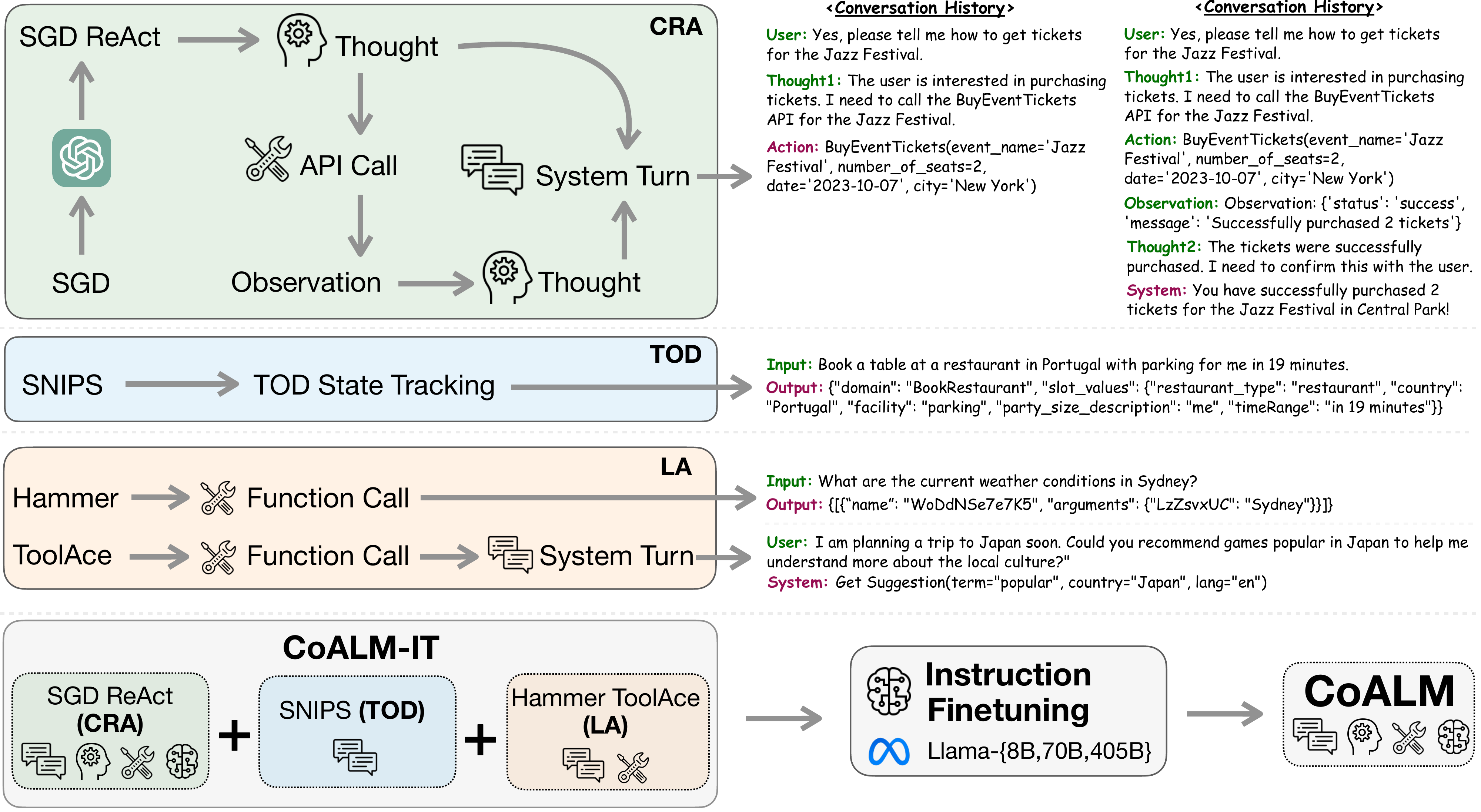}
\caption{\textbf{Overview of the CoALM Pipeline.} This figure illustrates our dataset generation and fine-tuning framework. The top three rows depict the data transformation processes, along with a corresponding sample shown on the right. In each training sample, \textbf{\textcolor{darkgreen}{green}} text highlights the input components of the instruction sample, while \textbf{\textcolor{mypurple}{purple}} text indicates the target outputs optimized during fine-tuning. For detailed examples, refer to Figures \ref{tab:snips-dst} - \ref{tab:sgd-sft-response}.}
\vspace{-2.0ex}
\label{fig:summary}
\end{figure*}

\textit{What if a single model could master both conversational and agentic tasks at the same time?}
The narrative of our paper aims to address the vision of a unified \textit{conversational agent}. 
Such an agent must excel not only in handling multi-turn conversations and TOD tasks but also in leveraging advanced LA capabilities, such as compound tool usage.
Previous research has focused on training dialogue agents in controlled scenarios (e.g., booking and reservation tasks) \cite{li2024largelanguagemodelszeroshot} with limited set of functions coming from dialogue actions (e.g., find\_attraction, book\_hotel), or, relied on hand-crafted long prompts tied to a small set of predefined APIs \cite{xu-etal-2024-rethinking} leveraged by propriety models like GPT-4. However, these approaches face limitations in real-world applications. Specifically, existing systems cannot easily adapt to new services without expensive fine-tuning or prompt engineering, yet real-world users often need access to a diverse range of APIs and functionalities according to their needs. 
Moreover, previous work shown a notable performance gap reported in TOD tasks between closed-source and open-source models \cite{hudecek-dusek-2023-large, xu-etal-2024-rethinking, li2024largelanguagemodelszeroshot}.
This tension underscores the need for an integrated open-source framework that supports both long-term dialogue state tracking and complex function calling from wide variety of APIs\footnote{In this work, words such as "tool use", "function calling", and "API calling" are used interchangeably.}.

We propose \textbf{CoALM} (Conversational Agentic Language Model), a unified approach that integrates TOD strengths (e.g., multi-turn state tracking) with LA capabilities (e.g., dynamic tool use). To achieve this, we develop \textbf{CoALM-IT}, a dataset spanning three dimensions: dialogue state tracking, complex function calling, and multi-turn conversations in ReAct style where the agent integrates its reasoning process with actions before providing the final response \cite{yao2023reactsynergizingreasoningacting-react}. The novelty of CoALM-IT comes from its Conversational ReAct API (CRA) instances, which makes it the first multi-turn TOD dataset explicitly incorporating ReAct-style reasoning with multiple think steps inside, generated using GPT-4o. The first think steps are responsible for deciding to call an API or not and second think step is to decide whether to response to user or not. Leveraging CoALM-IT, we trained CoALM model series: \textbf{CoALM 8B}, \textbf{CoALM 70B}, and \textbf{CoALM 405B}, a family of conversational agents demonstrates state-of-the-art performance on both TOD and LA domains. To comprehensively evaluate this, we perform experiments on one TOD benchmark, MultiWOZ 2.4 \cite{ye-etal-2022-multiwoz}, and two popular function calling benchmarks, the Berkeley Function Calling Leaderboard V3 (BFCL) \cite{bfcl} and API-Bank \cite{li-etal-2023-apibank} in completely zero-shot settings\footnote{Here, "zero-shot" refers to none of the evaluation benchmark train-set was used while training the CoALM models with CoALM-IT.}.

Our experiments reveal a stark gap in existing models: while LAs excel at tool calling on BFCL V3, they falter on MultiWOZ 2.4 with poor task completion. Conversely, base LLMs and traditional TOD systems show limited function calling abilities, as evidenced by the low performance on BFCL V3 and API-Bank. In contrast, our CoALM models, excel across both TOD and LA tasks. Our larger-scale open-source variants—CoALM 70B and CoALM 405B—outperform GPT-4o and other domain-specific models on both TOD (MultiWOZ) and function calling benchmarks (BFCL V3 and API-Bank).

In this paper, we study: \textit{How can we craft a single conversational agentic LLM that elegantly interweaves multi-turn dialogue mastery with powerful function calling capabilities?} Our key contributions are as follows:
\begin{itemize}[topsep=2pt, partopsep=-5pt, leftmargin=8pt, itemsep=-4.5pt]
    \item We analyze the gap between two domains: TOD systems and LA through evaluations on MultiWOZ 2.4, BFCL V3, and API-Bank, showing limitations of existing approaches.
    \item We introduce \textbf{CoALM-IT}, a hybrid multi-task dataset for conversational agents that, for the first time, explicitly incorporates ReAct-style reasoning steps in multi-turn TOD scenarios. Notably, to our knowledge, no prior effort has trained ReAct-based models using multi-turn TOD data in this manner.
    \item We propose \textbf{CoALM}, a family of model series trained with CoALM-IT: \textbf{CoALM 8B}, \textbf{CoALM 70B}, and the largest open-source conversational agent \textbf{CoALM 405B}—all unified by multi-turn dialogue skills and advanced function calling capabilities. 
     \item Our larger models, CoALM 70B and CoALM 405B, outperform GPT-4o and GPT-4o-mini on both TOD and LA tasks, narrowing gap between agents using closed-source and open-source models.
\end{itemize}
To foster further research within the open-source community, we publicly release code, all model weights, datasets, intermediate checkpoints, and training configurations.
\section{Related Work}

\noindent\textbf{Dialogues and the Domain Shift.} Earlier studies work on applying LLMs to dialog applications through supervised fine-tuning \cite{su-etal-2022-multi, gupta-etal-2022-instructdial} or different prompting methods \cite{hu-etal-2022-context, chung-etal-2023-instructtods, zhang-etal-2023-sgp}.
Following these, \citet{hudecek-dusek-2023-large} have examined the dialogue management abilities of instruction-tuned LLMs in handling goal-oriented multi-turn conversations.
More recently, existing work in dialogue agents primarily focuses on leveraging dialogue acts to derive API calls for backend services \cite{li2024largelanguagemodelszeroshot, xu-etal-2024-rethinking, king-flanigan-2024-unsupervised}.
FNCTOD \cite{li2024largelanguagemodelszeroshot} fine-tunes on a small dataset restricted to a limited set of domain-specific APIs for state tracking, whereas AutoTOD \cite{xu-etal-2024-rethinking} uses GPT-4 with hand-crafted prompts that rely on a narrow set of predefined APIs with long instructions for each dialogue domain. However, these approaches are brittle and difficult to scale in real life scenarios, as they require costly re-trainings or extensive prompt engineering to handle new services, unseen domains, and unexpected user requests. Our work aligns with these studies in building such agents, but CoALM can manage thousands of complex APIs at the same time and can generalize to unseen domains without expensive training cycles and time-intensive prompt engineering. 

\vspace{3mm}

\noindent\textbf{Language Agents.} Tool learning with LLMs has evolved from simple simple reasoning \cite{NEURIPS2022_9d560961-cot} to more sophisticated approaches \cite{yao2023reactsynergizingreasoningacting-react, tool_learning}. Early work relied on prompting to enable tool usage \cite{yao2023reactsynergizingreasoningacting-react, paranjape2023art}, but more recent research has focused on specialized fine-tuning approaches for effective function calling accuracy~\cite{schick2024toolformer, patil2023gorilla, wang2024executable, zhang2024xlam}. For example, Toolformer \cite{schick2024toolformer} have explored how LLMs autonomously learn when and how to call APIs, leading to improved performance in task-specific settings. In this direction, recent works \cite{abdelaziz-etal-2024-granite, Liu2024ToolACE, Lin2024Hammer} focus on fine-tuning synthetically generated data to integrate more complex tool calling capabilities, such as nested function calls and irrelevance detection. These approaches shown promising results on LA benchmarks, however they mostly operate on single-turn interactions with the user and fall short of enabling user-driven, multi-domain, and multi-turn task completion which is essential for real-world conversational systems.

\begin{table*}[!t]
\centering
\resizebox{\textwidth}{!}{
\begin{tabular}{lrrrrrr}
\toprule
\textbf{Data Domain} & \textbf{Data Type} & \textbf{Data Name} & \textbf{Data Format} & \textbf{\# of Data Samples} &  \textbf{\# of Total Tokens} & \textbf{Avg. Tokens Per Sample} \\ \midrule

\multirow{1}{*}{\textbf{TOD}}&  Single-Turn  & SNIPS         & State Tracking &    $13,028$   &   $12,278,780$ &  $942.49$  \\ \cmidrule{2-7}
\multirow{2}{*}{\textbf{LA}}& Single-Turn & Hammer  & API Call          &     $13,819$  &  $10,199,147$  &   $738.05$ \\
                                  & Multi-Turn  & ToolAce & API Call          &     $202,500$ &  $129,001,612$ &  $637.04$ \\ \cmidrule{2-7}
\multirow{1}{*}{\textbf{CRA}}& Multi-Turn & SGD & ReAct API Call &    $82,236$   &   $59,704,782$ &  $726.02$ \\ \cmidrule{2-7}

\multicolumn{4}{r}{\textbf{Total}} & $311,583$ &  $211,184,321$ &  $760.90$ \\ \bottomrule
\end{tabular}
}
\caption{\textbf{CoALM-IT Dataset Details.} Statistical details of our proposed CoALM-IT dataset showcasing the training mixtures. Generated \textbf{CRA} denotes the Conversational ReAct API dataset.}
\vspace{-5mm}
\label{tab: data_training_mixture_stats}
\end{table*}


                             



\section{Preliminaries}
A Conversational Agent, at its core, must understand user intents, maintain context across multi-turn interactions, and respond contextually. Beyond traditional TOD tasks, modern conversational agents are also expected to exhibit agentic abilities, like tool calling, planning, and decision making, to fulfill complex user requests. An effective conversational agent integrates these capabilities as skills, ensuring natural and relevant interactions while efficiently completing the user’s objectives. The detailed task formulations for TOD systems and LA are provided in Appendix \ref{app: problem-formulation}.

\vspace{-1ex}

\subsection{Why we need both TOD and LA Capabilities?}

Multi-turn interactions are critical for refining ambiguous user requests. For example, when a user says "Find me a hotel", the system can ask clarifying questions to clarify the user's intention (e.g., location, price range) instead of returning generic results. This ensures meaningful and task-specific conversations. That said, traditional TOD systems excel at handling these multi-turn interactions but over a small set of APIs (e.g., query\_restaurant, book\_hotel) \cite{ye-etal-2022-multiwoz}. By training on structured dialogue flows, they achieve high task success rates in controlled scenarios (e.g., standard booking or reservation tasks) without requiring complex function-calling capabilities. However, these systems struggle to adapt to new services (e.g., airline, retail) without expensive re-training. 

In real-world settings, users may need to access a wide variety of APIs (e.g., search\_direct\_flight, get\_product\_details). This is where LA shines: they leverage LLMs and can rapidly learn how to use unseen new tools since they are already proficient with determining when to invoke an API and decide which API to use from a diverse set of available functions. Without these skills, agents fail to fulfill complex user goals, limiting their utility. 

Together, these skills form the backbone of a unified conversational agents, enabling them to transition from being passive responders to proactive collaborators capable of managing intricate tasks and sustaining user engagement.

\subsection{Can TOD Systems Solve Function Calling Tasks?}
The benchmark results demonstrate the limitations of TOD systems in function calling scenarios. Despite achieving top performance on MultiWOZ metrics as in Table \ref{tab:tod-results}, these systems show significantly lower accuracy on both API-Bank (Table \ref{tab:apibank}) and BFCL (Table \ref{tab:bfcl}) benchmarks. This performance gap reveals that TOD systems' traditional strengths in dialogue management do not translate well to handling diverse, unseen, and complex API calls.

\subsection{Can LAs Handle Task-oriented Multi-turn Conversations?}
Conversely, agentic models like ToolAce \cite{Liu2024ToolACE}, Hammer \cite{Lin2024Hammer}, and Granite \cite{abdelaziz-etal-2024-granite} while achieving accurate results on API-Bank and BFCL V3, perform poorly on MultiWOZ's task completion metrics. These results highlight a critical weakness: while they deliver strong performance on function execution tasks, they fall short in maintaining coherent multi-turn conversations and properly fulfilling user intents. Their specialized optimization for tool calling impairs their dialogue management abilities, indicating that current LAs need more balanced capabilities to handle task-oriented conversations more effectively.

\section{Methodology}
Our approach, illustrated in Figure \ref{fig:summary}, develops a unified agent skilled in goal-oriented multi-turn conversations and function calling. First, we build the CoALM-IT, a broad instruction-tuning (IT) dataset that spans multiple domains, tasks, and unique reasoning structures. Next, we do fine-tuning on the proposed CoALM-IT dataset to produce CoALM; a balanced conversational agent model series capable of complex reasoning, fluent dialogue, user intent fulfillment, and function calling.

\subsection{Conversational Agent Dataset Generation}
\label{sec:dataset-generation}
To develop a conversational agent with diverse capabilities, we created a comprehensive dataset that combines samples across multiple skills essential for both multi-turn task-oriented conversations and tool utilization. Figure \ref{fig:summary} summarizes how the dataset is created and Table \ref{tab: data_training_mixture_stats} provides detailed statistics of CoALM-IT.

\vspace{3mm}

\noindent\textbf{TOD Datasets.} An accurate dialogue system needs to master three fundamental capabilities: providing accurate information to users, fulfilling user goals, and tracking dialogue states to understand user intents and goals throughout conversations \cite{paradise}. To equip our model with these skills, we utilized the SNIPS dataset \cite{coucke2018snips}, originally designed for language understanding but repurposed for single-turn dialogue state tracking (DST). We extracted its training split and converted it into the state tracking IT format by crafting a detailed instruction prompt, as illustrated in Figure \ref{tab:snips-dst}. This transformation resulted in a training set of 24,542 samples for effective DST.

\vspace{3mm}

\noindent\textbf{Function Calling Datasets.} Tool calling capability is the ability to select appropriate APIs and access external knowledge, which is crucial in modern LAs. An effective agent must not only choose the correct API but also provide properly typed parameters (e.g., integers or strings) and manage complex scenarios involving sequential or parallel function calls. To develop these skills, we incorporated datasets from two state-of-the-art agent models: Hammer \cite{Lin2024Hammer} and ToolACE \cite{Liu2024ToolACE}. Hammer's training dataset incorporates random noise by replacing function and parameter names to prevent overfitting (see Figure \ref{fig:summary}), forcing the model to reason about API functionality through provided descriptions rather than memorizing specific identifiers. ToolACE provides multi-turn conversational scenarios in open-domain settings, where function calls may occur across multiple turns, but no database is provided. We post-process these datasets by incorporating the prompt instructions and adding conversation history if available. As reported in Table \ref{tab: data_training_mixture_stats}, the combined API calling corpus contains 216,319 samples. A function calling training sample for the Hammer dataset can be seen in Figure \ref{tab:hammer-api-2}.  

\vspace{3mm}

\noindent\textbf{Conversational ReAct-based API Calling (CRA) Dataset.} While state tracking enables the understanding of user intent and function calling provides external knowledge access, integrating these capabilities within multi-turn task-oriented conversations requires additional reasoning about when to make API calls and how to interpret their results. Our primary contribution is a completely new User and Agent conversation structure as \textbf{User-Thought1-Action-Observation-Thought2-Response}. Starting from multi-turn SGD dataset \cite{rastogi2020sgd}, we systematically transform each turn to include two distinct reasoning steps (Thought1 and Thought2) and potential API calls (Action and Observation), extending traditional ReAct format \cite{yao2023reactsynergizingreasoningacting-react} by incorporating GPT-4o for content generation (Figure \ref{fig:summary} top row). Our structure includes two main parts: (i) \textbf{User-Thought1-Action}, which focuses on understanding the user's intent with reasoning and invoking the right API, if necessary (Figure \ref{tab:sgd-sft-action} bottom). (ii) \textbf{Observation-Thought2-Response}, where the agent analyzes the returned observations and formulates an appropriate response to the user (Figure \ref{tab:sgd-sft-response} bottom). This transformation is achieved with a carefully designed prompt in Table \ref{tab: prompt}, which enforces strict “Role Definition”, “Task Information”, and “Output Format”. Since CRA is generated via GPT-4o \cite{Achiam2023GPT4TR}, it is also validated by human evaluators (Appendix \ref{human-validation}). Best of our knowledge, this is the first ReAct-based Conversational API dataset that incorporates multiple intermediate reasoning steps in multi-turn settings for TOD. This process yielded 82,236 samples, specifically tailored for task-oriented domains such as hotel bookings and restaurant reservations. 

\vspace{3mm}

We merge all three datasets into a single training set called CoALM-IT, please refer to Table \ref{tab: data_training_mixture_stats} for details. We fine-tune our CoALM models on this merged dataset in one pass. By interleaving samples from TOD, LA, and CRA, the model continuously practices different conversational skills without overfitting to any single domain or task type.

\subsection{Fine-tuning Towards Conversational Agents}
We followed a multitask fine-tuning approach to develop CoALM models' diverse capabilities across TOD, function calling, and multi-turn reasoning by training on CoALM-IT. Our training process is structured to target specific skills through different optimization objectives \textbf{completely in zero-shot settings}, as our CoALM-IT dataset does not contain any of the evaluation benchmark training sets.

\vspace{3mm}

\noindent\textbf{Multitask Fine-tuning.}  As described in Section \ref{sec:dataset-generation} and illustrated in Figure \ref{fig:summary}, our CoALM-IT dataset combines samples from three distinct domains, each designed to cultivate a specific skill: (i) TOD (Task-Oriented Dialogue) for strengthening dialogue state tracking, (ii) LA (Language Agent) for teaching the model when and how to invoke function calls, and (iii) ReAct for multi-turn conversation, multi-step reasoning and function calling.

For TOD, we augment SNIPS data with prompt instructions (Figure \ref{tab:snips-dst}), training the model to generate structured dialogue states in response to user queries. For function calling (LA), we optimize CoALM to select the correct APIs and produce accurate function calls with proper parameter types (Figure \ref{tab:hammer-api-2}), emphasizing reasoning over memorized patterns. We then address complex multi-turn conversations with API integration using our CRA dataset, formatted in the ReAct style. This stage uses two objectives: (1) action prediction (Figure \ref{tab:sgd-sft-action}), where the model learns to issue the appropriate function call given the conversation history, and (2) response generation (Figure \ref{tab:sgd-sft-response}), where it synthesizes coherent replies based on both API results and intermediate reasoning steps. Rather than merely producing answers, the model learns to reason, decide, and act in multiple stages before arriving at a final response. Notably, we trained our models on CoALM-IT by interleaving TOD, LA, and CRA samples, enabling the model to continuously practice diverse conversational skills while avoiding overfitting to any single domain or task type.

\vspace{3mm}

\noindent\textbf{Training Details.} We developed the CoALM model series by fine-tuning Llama 3.1 8B, Llama 3.3 70B, and Llama 3.1 405B \cite{Dubey2024TheL3-llama3} using a consistent Alpaca (Instruction-Input-Output) format. To balance efficiency and model quality, we applied LoRA \cite{hu2021lora} rank (r) = 16 and scaling factor ($\alpha$) = 32 to all linear layers, and trained in mixed-precision bfloat16 (bf16) on 8 NVIDIA H100 GPUs. Under these settings, CoALM 8B required approximately 8 hours of training, while CoALM 70B took about 60 hours. We used a global batch size of 8, trained for 3 epochs with a learning rate of $1e-4$, and employed a linear warm-up schedule with a 0.1 ratio. For CoALM 405B, we fine-tuned Llama 3.1 405B and using QLoRA \cite{dettmers2023qloraefficientfinetuningquantized} with the same rank and scaling factor using bitsandbytes \cite{bitsandbytes} with a quantization type of normalized float 4 (nf4). The precise  training configurations for CoALM 8B, CoALM 70B and CoALM 405 are included in the HuggingFace pages. Our training pipeline leveraged the Oumi framework\footnote{\url{https://github.com/oumi-ai/oumi}} to ensure reproducibility and streamlined management \cite{oumi2025}.

\section{Experiments}
This section presents results highlighting CoALM's effectiveness in unifying conversational management and advanced API calling, outperforming specialized models across both TOD and LA benchmarks.

\begin{table}[!t]
\centering
\resizebox{1.0\columnwidth}{!}{
\begin{tabular}{l c c}
\toprule
\textbf{Method}                                   & \textbf{Success} & \textbf{JGA}    \\ \midrule
CoALM 8B (ours)                                    & 51.6             & 30.4            \\
CoALM 70B (ours)                                   & 69.4             & \textbf{43.8}   \\
CoALM 405B (ours)\textbf{$^*$}                                & 66.7             & 38.8             \\ \midrule
Hammer 2.0 7B                                     & 23.5             & 21.7            \\
ToolAce                                           & 18.0             & 34.4            \\ 
Granite-20B-Code                                  & 10.7             & 21.8            \\ 
CodeActAgent                                      & 9.5              & 20.2             \\ 
Llama 3.1 8B Instruct                             & 19.9             & 26.3             \\ 
Llama 3.3 70B Instruct                            & 67.6             & 40.8             \\ 
Mistral-7B-Instruct-v0.3                          & 31.2             & 27.0             \\ 
FNCTOD \cite{li2024largelanguagemodelszeroshot}   & 44.4             & 37.9             \\
NC-Latent-TOD \cite{king-flanigan-2024-unsupervised} & 68.3          & 39.7             \\
GPT 3.5 Turbo \cite{hudecek-dusek-2023-large}     & -                & 13.5             \\
GPT4o-mini                                        & 69.9             & 38.4             \\
GPT4o                                             & \textbf{75.5}             & 36.9             \\ \bottomrule
\end{tabular}
}
\caption{\textbf{MultiWOZ 2.4 Benchmark Results.} Performance comparison across models on MultiWOZ 2.4 dialogue benchmark. Best scores are highlighted with \textbf{bold}. The asterisk (*) on CoALM 405B denotes the checkpoint from one completed epoch, as the model is still under training.}
\vspace{-5mm}
\label{tab:tod-results}
\end{table}

\subsection{Experimental Settings}

\noindent\textbf{Evaluation Benchmarks.} We evaluate our approach on three complementary benchmarks that assess different aspects of model performance: MultiWOZ 2.4 (TOD), API-Bank (LA), and BFCL V3 (LA). Specifically, MultiWOZ 2.4~\cite{ye-etal-2022-multiwoz} is a multi-domain TOD dataset covering scenarios such as hotel booking and transportation, where we measure Success Rate and Joint Goal Accuracy (JGA); in our zero-shot setting, we rely on the test set of 999 samples, using a slightly modified AutoTOD prompt~\cite{xu-etal-2024-rethinking}. API-Bank~\cite{li-etal-2023-apibank} focuses on evaluating tool-augmented LAs through 314 tool-use dialogues and 753 API calls, tested at two levels: L-1 (invoking a known API) and L-2 (retrieving and calling from multiple candidates). Lastly, BFCL V3\footnote{\url{https://gorilla.cs.berkeley.edu/blogs/13_bfcl_v3_multi_turn.html}}~\cite{patil2023gorilla} provides over 1{,}800 test cases spanning tasks like simple, multiple, and parallel function calls, evaluated by Abstract Syntax Tree (AST) accuracy and Executable Function Accuracy. See Appendix~\ref{app:benchmark-details} for further details. 

\vspace{3mm}

\noindent\textbf{Baselines.} In the LA tasks, we included strong baselines like Hammer \cite{Lin2024Hammer}, ToolAce \cite{Liu2024ToolACE}, Granite \cite{abdelaziz-etal-2024-granite} which represent state-of-the-art models in agentic tasks, including OpenAI models. For MultiWOZ evaluations, we recognize that many existing TOD models are trained with classification-based supervised fine-tuning, focusing primarily on DST. Such models do not support free-form dialogue generation, nor do they exhibit broader “chat” capabilities. In contrast, our approach aims to unify both conversational (LA) and agentic (TOD) tasks into a single, generative framework. On the other hand, there are some models evaluated in zero-shot settings but as per domain JGA, rather than overall JGA. That said, we used top popular zero-shot models FNCTOD \cite{li2024largelanguagemodelszeroshot} and NC-Latent-TOD \cite{king-flanigan-2024-unsupervised} as our TOD baselines in TOD. Please see Appendix \ref{app: model-overview} for more details of these baseline models.

\begin{table}[!t]
\centering
\resizebox{1.0\columnwidth}{!}{
\begin{tabular}{lrrrrrrrr}
\toprule
\multicolumn{1}{c}{\multirow{2}{*}{\textbf{Model}}} & \multicolumn{2}{c}{\textbf{Rouge-L*}} & \multicolumn{2}{c}{\textbf{Rouge-1}} & \multicolumn{2}{c}{\textbf{Rouge-2}} & \multicolumn{2}{c}{\textbf{BLEU-4}} \\
\multicolumn{1}{c}{} &
  \multicolumn{1}{c}{L-1} &
  \multicolumn{1}{c}{L-2} &
  \multicolumn{1}{c}{L-1} &
  \multicolumn{1}{c}{L-2} &
  \multicolumn{1}{c}{L-1} &
  \multicolumn{1}{c}{L-2} &
  \multicolumn{1}{c}{L-1} &
  \multicolumn{1}{c}{L-2} \\ \midrule
  
CoALM 8B (ours)               & 92.8 & \underline{81.9} & 94.1 & 81.2 & 91.9 & \underline{76.4} & 89.4 & \underline{69.7} \\
CoALM 70B (ours)              & \underline{92.7} & \textbf{83.2}  & \underline{94.5} & \textbf{82.7}  & \textbf{92.5}  & \textbf{78.9}  & \underline{89.5}  & \textbf{72.4}  \\
CoALM 405B (ours)\textbf{$^*$}             & \textbf{93.4} & 77.8  & \textbf{94.5} & 77.1  & \underline{92.4}  & 71.9  & \textbf{90.3}  & 64.4  \\ \midrule

Llama 3.1 8B Instruct  & 72.7 & 75.2 & 84.0 & \underline{81.4} & 79.8 & 76.3 & 62.3 & 65.1 \\
Qwen2.5 7B Instruct    & 84.3 & 73.9 & 88.9 & 78.5 & 84.6 & 71.2 & 76.4 & 64.2 \\

Hammer 2.0 7B          & 90.1 & 74.0 & 92.3 & 74.1          & 89.9 & 68.5 & 85.4 & 58.4 \\ 
ToolAce                & 81.5 & 63.6 & 88.8 & 71.3          & 85.0 & 63.0 & 76.1 & 67.0 \\
Granite-20B-Code       & 60.3 & 45.7 & 64.7 & 48.9          & 59.5 & 43.4 & 43.8 & 29.3 \\\midrule

Fnc-TOD 13B            & 3.9 & 3.3 & 22.1 & 23.4          & 8.0 & 9.2 & 1.5 & 1.1 \\
LDST                   & 8.3 & 7.1 & 12.8 & 11.6          & 2.7 & 2.4 & 6.2 & 5.7 \\
tod-zero-bqag3oyb      & 3.7 & 4.2 & 11.5 & 12.4          & 1.1 & 2.2 & 1.0 & 0.9 \\
nc-latent-tod-step-2   & 3.2 & 3.2 & 14.3 & 13.3          & 3.2 & 1.5 & 0.8 & 0.8 \\ \bottomrule
\end{tabular}
}
\caption{\textbf{API-Bank Benchmark Results.} Performance comparison across models on API-Bank function calling benchmark. Best scores are highlighted with \textbf{bold} and the second-best results are \underline{underlined}. The asterisk (*) on CoALM 405B denotes one completed epoch, as the model is still in the training process.}
\vspace{-5mm}
\label{tab:apibank}
\end{table}

\begin{table*}[!t]
\centering
\resizebox{1.0\linewidth}{!}{
\begin{tabular}{l c c c c c c c }
\toprule
\textbf{Model}                  & \textbf{Overall Acc}   & \textbf{Non-Live AST Acc} &  \textbf{Non-Live Exec Acc} & \textbf{Live Acc}    & \textbf{Multi Turn Acc}  & \textbf{Relevance Detection} & \textbf{Irrelevance Detection}  \\ \midrule

Mistral-7B-Instruct-v0.3        & 38.35\%                & 56.33\%                   & 63.77\%                     & 57.31\%              & 0.25\%                   & 77.78\%                      & 41.84\%                         \\
Llama-3.1-8B-Instruct           & 49.84\%                & 84.25\%                   & 79.75\%                     & 60.33\%              & 10.25\%                  & 75.61\%                      & 47.92\%                         \\
Llama-3.3-70B-Instruct          & 51.36\%                & 84.85\%                   & \underline{90.05\%}         & 62.51\%              & 7.25\%                   & \underline{95.12\%}          & 48.33\%                         \\
ToolAce                         & 52.55\%                & 82.19\%                   & 86.98\%                     & 71.08\%              & 0.88\%                   & 70.73\%                      & 87.29\%                        \\
Hammer2.0-7b                    & 52.13\%                & 86.94\%                   & 83.66\%                     & 71.17\%              & 0.38\%                   & 95.12\%                      & 73.20\%                         \\
Llama-3.1-405B-Instruct         & 56.38\%                & \underline{89.71\%}       & 84.70\%                     & 70.77\%              & 11.75\%                  & 88.89\%                      & 70.86\%                        \\
GPT-4o-mini (2024-07-18)        & 59.40\%                & 86.52\%                   & \textbf{85.05\%}            & 73.26\%              & 19.00\%                  & 78.05\%                      & 76.97\%                         \\
GPT-4o (2024-08-06)             & 59.83\%                & 70.08\%                   & 60.79\%                     & \textbf{76.41\%}     & \textbf{34.62\%}         & 51.22\%                      & \textbf{87.34\%}                \\ \midrule

CoALM 8B (ours)                  & 54.11\%                & 85.17\%                   & 78.61\%                     & 72.59\%              & 7.00\%                   & 77.78\%                      & 83.00\%                        \\ 
CoALM 70B (ours)                 & \underline{60.49\%}    & 82.94\%                   & 81.36\%                     & 72.19\%              & 26.25\%                  & 72.22\%                      & \underline{85.36\%}            \\ 
CoALM 405B (ours)\textbf{$^*$}                & \textbf{63.34\%}       & \textbf{90.46\%}          & 84.75\%                     & \underline{74.59\%}  & \underline{28.25\%}      & \textbf{100.00\%}            & 72.26\%                        \\ 
 \bottomrule
\end{tabular}
}
\caption{\textbf{BFCL V3 Benchmark Results.} Performance comparison on the BFCL V3 function-calling benchmark. The best results are highlighted in \textbf{bold}, while the second-best results are \underline{underlined}. The asterisk (*) on CoALM 405B denotes one completed epoch, as the model continues training.}
\label{tab:bfcl}
\end{table*}

\begin{table*}[!t]
\centering
\resizebox{1.0\linewidth}{!}{
\begin{tabular}{l cc c cc c c}
\toprule
& \multicolumn{2}{c}{\textbf{TOD Task}} & & \multicolumn{4}{c}{\textbf{Function Calling Tasks}}  \\ \cmidrule{2-3} \cmidrule{5-8} 
& \multicolumn{2}{c}{\textbf{MultiWOZ 2.4}} & & \multicolumn{2}{c}{\textbf{API-Bank}}               & & \multicolumn{1}{c}{\textbf{BFCL-V3}} \\
  \cmidrule{2-3}
  \cmidrule{5-6}
  \cmidrule{8-8}
 Model & Success & DST &  & Rouge-L1 & Rouge-L2&  & Overall Success \\ \midrule
\textcolor{darkgreen}{}
Llama 3.1 8B Instruct        & 19.9  & 26.3  &  & 72.7  & 75.2 &  &49.8    \\
$\;\;$ + CoALM-IT w/o LA     & 46.0 \small{($\textcolor{darkgreen}{26.1}\uparrow, \textcolor{red}{5.6}\downarrow$)}   \textcolor{red} & 28.5 \small{($\textcolor{darkgreen}{2.2}\uparrow, \textcolor{red}{1.9}\downarrow$)}  &  & 45.5 \small{($\textcolor{red}{27.2}\downarrow, \textcolor{red}{47.3}\downarrow$)} & 48.8 \small{($\textcolor{red}{26.4}\downarrow, \textcolor{red}{33.1}\downarrow$)} &  & 35.4 \small{($\textcolor{red}{14.4}\downarrow, \textcolor{red}{18.3}\downarrow$)}   \\
$\;\;$ + CoALM-IT w/o TOD   & 42.0 \small{($\textcolor{darkgreen}{22.1}\uparrow, \textcolor{red}{9.6}\downarrow$)}  & 19.4 \small{($\textcolor{red}{6.9}\downarrow, \textcolor{red}{11.0}\downarrow$)}  &  & 92.7 \small{($\textcolor{darkgreen}{20.0}\uparrow, \textcolor{red}{0.1}\downarrow$)}     & 78.9 \small{($\textcolor{darkgreen}{13.7}\uparrow, \textcolor{red}{3.0}\downarrow$)} &  & 54.1 \small{($\textcolor{darkgreen}{4.3}\uparrow, \textcolor{darkgreen}{0.4}\uparrow$)}   \\
$\;\;$ + CoALM-IT w/o CRA    & 50.0 \small{($\textcolor{darkgreen}{30.1}\uparrow, \textcolor{red}{1.6}\downarrow$)}  & \textbf{34.5} \small{($\textcolor{darkgreen}{8.2}\uparrow, \textcolor{darkgreen}{4.1}\uparrow$)}  &  & 91.3 \small{($\textcolor{darkgreen}{18.6}\uparrow, \textcolor{red}{1.5}\downarrow$)}  & 78.8 \small{($\textcolor{darkgreen}{3.6}\uparrow, \textcolor{red}{3.1}\downarrow$)} &  & \textbf{56.6} \small{($\textcolor{darkgreen}{10.6}\uparrow, \textcolor{darkgreen}{2.9}\uparrow$)}      \\ \midrule
CoALM 8B                         & \textbf{51.6} & 30.4  &  & \textbf{92.8}  & \textbf{81.9} &  & 53.7   \\ \bottomrule

\end{tabular}
}

\caption{\textbf{Dataset Domain Effects.}
Experimental results highlighting the impact of excluding specific domain datasets during CoALM fine-tuning. \textbf{w/o} indicates excluding the corresponding dataset during fine-tuning. Each row displays performance changes in parentheses with respect to base model (Llama) and final model (CoALM), i.e. ($\Delta$ Llama, $\Delta$ CoALM). Performance gains are highlighted in \textbf{\textcolor{darkgreen}{green}}, while drops are marked in \textbf{\textcolor{red}{red}}.
}
\vspace{-1mm}
\label{tab:dataset-domain}
\end{table*}

\subsection{Results on MultiWOZ}
\vspace{3mm}

\noindent\textbf{LA models struggle with TOD.} Table~\ref{tab:tod-results} summarizes results on MultiWOZ 2.4. Baseline models optimized for function calling (ToolAce, Hammer, Granite, CodeAct) achieve low Success Rate and JGA. Although these agents can call APIs effectively, they fail to track user intents across multiple sessions or deliver correct final answers to the user, except ToolAce JGA reaches 34.4\% accuracy close with domain-specific TOD models like FNCTOD.  Instruction-tuned base LLMs like Llama 3.1 8B perform moderately better on MultiWOZ, reaching a 19.9\% Success rate and 26.3\% JGA. 

\vspace{3mm}

\noindent\textbf{CoALM surpasses and generalizes in TOD.} In contrast, our smallest CoALM 8B achieves 51.6\% Success Rate, more than doubling the Success performance compared to Llama 3.1 8B and surpassing other LAs. Moreover, our CoALM 70B model achieves top results on DST with achieving 43.8\% JGA, even outperforming GPT-4o and GPT-4o-mini. This shows CoALM’s ability with coherent multi-turn state-tracking, outperforming existing baselines and domain-specific models like FNCTOD.  Notably, CoALM's strong performance is achieved without any MultiWOZ samples in its CoALM-IT training dataset, demonstrating its robustness in out-of-distribution (OOD) generalization.

\subsection{Results on API-Bank and BFCL}

\noindent\textbf{CoALM adeptly orchestrates function calls.} Table~\ref{tab:apibank} shows API-Bank scores to test model's API calling capabilities where Rouge-L is the primary evaluation metric. TOD models in the bottom row yield suboptimal results in this task. On the other hand, CoALM 8B achieves a Rouge-L score of 92.8 at Level-1 and 81.9 at Level-2, surpassing both TOD-oriented models and tool-centric LAs by a significant margin. It also achieves top performance on nearly all metrics. Moreover, we scale CoALM 8B accuracy with CoALM 70B and CoALM 405B models achieving top best and second best scores. This suggests that CoALM’s balanced approach enables it not only to retrieve and call the correct API but also to generate precise responses grounded in the returned results, fulfilling complex user requests effectively. 

\vspace{3mm}

\noindent\textbf{CoALM outperforms specialized LAs and GPT-4o.} We next assess function calling accuracy on BFCL V3 (Table~\ref{tab:bfcl}). Models trained only for TOD or basic instruction-following underperform. While LAs like Hammer and ToolAce fare better, our smallest model CoALM 8B surpasses them (see Figure \ref{fig:error-analysis} for error analysis examples). Our larger scale models outperform GPT-4o, GPT-4o-mini and Llama-3.1-405B in overall accuracy. Remarkably, CoALM 405B achieves 100\% accuracy on the relevance detection task, highlighting its agentic reasoning capabilities through hallucination. CoALM 405B stands as the top-performing fully open-source model on BFCL V3 leaderboard.

\vspace{3mm}

\subsection{Domain Impact on Performance}
Table~\ref{tab:dataset-domain} highlights the performance impact of CoALM-IT's fine-tuning components. Removing LA datasets significantly reduces function calling performance, with API-Bank Rouge-L1 dropping 47.3\% and BFCL success falling 18.3\%.
Excluding the DST dataset leads to a notable decline in CoALM's JGA, dropping by 11.0\% relative to CoALM and even underperforming base Llama by 6.9\%. This underscores the essential role of fine-tuning on state tracking to capture user intents effectively.
Finally, removing the GPT-4-generated CRA dataset has negative impact on MultiWOZ 2.4's Success metric, which plummets by 11.7\%. Also, multi-turn function calling accuracy dropped in API-Bank, both in L1 and L2 metrics. This indicates that the CRA dataset is instrumental in developing coherent and contextually aware responses in multi-turn settings. However, JGA and BFCL's overall success see slight improvements, suggesting that certain specialized skills may benefit marginally in the absence of broader conversational reasoning.
These results confirm that each dataset is crucial for balanced task performance, enabling CoALM to generalize effectively across different tasks without overfitting to one domain.
\section{Conclusion and Future Work}
In this work, we highlighted a critical gap between LA and TOD systems, where each excels in complementary capabilities - function calling and multi-turn conversation management, respectively. To solve this, we introduced CoALM, unified conversational agents that seamlessly integrates sophisticated API usage with natural multi-turn dialogue. 
Through fine-tuning on CoALM-IT with a hybrid fine-tuning strategy, CoALM achieves leading performance on both TOD and LA benchmarks, demonstrating that a single model can indeed master multi-turn conversations and tool use effectively. 

Future work can investigate using reinforcement learning (RL) to generate large-scale interaction trajectories supported with API calls could further enhance the self-evolution of conversational agents through purely RL-based optimization. 
Another direction is, improving multi-turn function calling and user interaction abilities of these models, which remains a difficult problem with generally low accuracy. 
We believe that our findings, methodologies, and published resources will foster future research to create more capable and versatile conversational systems.

\section{Limitations}
While CoALM demonstrates improved performance across both conversational TOD and agentic tasks, we conducted all experiments solely using the Llama model family, limiting our insights into other architectures like Mistral and Qwen.
Furthermore, many TOD systems rely on classification-based supervised fine-tuning (DST-only), lacking free-form chat capabilities, so we are not able to integrate them in our chat-based evaluation setup for head-to-head comparisons.
We also did not systematically assess CoALM’s general reasoning abilities after post-training, leaving open the question of potential catastrophic forgetting if any.
Even though we introduced the open source model CoALM 405B, the computational cost of doing inference with CoALM 405B requires 16 H100 GPUs, which may limit accessibility for some researchers. 
Lastly, our current approach still relies on curated fine-tuning data; future work might investigate self-evolving methods that learns complex function calling skills continuously leveraging RL.
\section{Acknowledgements}
We would like to acknowledge the Oumi AI team \cite{oumi2025} for their assistance in training and scaling with the larger CoALM models. We would also like to thank Together AI \cite{togetherai} for providing the cluster resources necessary to enable CoALM 405B training. This project also has benefited from the Microsoft Accelerate Foundation
Models Research (AFMR) grant program, through which leading foundation models hosted by Microsoft Azure and access to Azure credits were provided to conduct the research.



\bibliography{custom}

\newpage
\clearpage
\appendix

\section*{Appendix}
\label{sec:appendix}

\section{Problem Formulation}
\label{app: problem-formulation}

\subsection{End-to-End TOD Systems with LLMs} LLM-based end-to-end TOD systems generate contextually relevant responses based on dialogue history and task instructions. Let $F$ be a language model parameterized by $\theta$, which maps an input context given as prompt $T$ to an output system response $y_t$. At each dialogue turn $t$, the system receives three key components: task instructions $G$, dialogue history $H_t$ comprising of prior user-system interactions $\{(u_1, y_1), ..., (u_{t-1}, y_{t-1})\}$, and the current user input $u_t$. These elements are combined to form the complete prompt $T_t = (G, H_t, u_t)$. The model generates a response $y_t$ by modeling the conditional probability:
\begin{equation}
    P(y_t|T_t; \theta) = P(y_t|G, H_t, u_t; \theta),
\end{equation}
where $P(s_t|T_t; \theta)$ denotes the probability of generating the response $y_t$ given the prompt $T_t$ and the model parameters $\theta$. The dialogue progresses by updating the history after each turn $H_{t+1} = H_t + {[(u_t, s_t)]}$, maintaining the sequential nature of the interaction while preserving task orientation through $G$.

\subsection{Function Calling with Language Agents} A language model $F_\theta$ maps an input $x = (G, u, \Omega)$, where $G$ is the task prompt, $u$ is the user query, and $\Omega = \{f_1, \dots, f_n\}$ is the set of available functions with their arguments and descriptions to a structured function call $y$. The model generates target function call in a structured format, such as JSON or text schema. The generation probability is defined as:
\begin{equation}
P(y \mid x; \theta) = P(y \mid G, u, \Omega; \theta)
\end{equation}
This formulation enables the model to translate natural language inputs into precise and well-structured function calls, facilitating seamless integration with external systems.

\paragraph{ReAct Prompting.} ReAct \cite{yao2023reactsynergizingreasoningacting-react} integrate reasoning and action-taking to enable more effective decision-making. It facilitates intermediate reasoning by breaking down complex tasks into smaller, interpretable reasoning steps. Additionally, it enables interaction with external tools or APIs by producing structured actions that integrate effectively with external systems. As a result of an API execution, ReAct incorporates observations dynamically, adapting subsequent reasoning and actions based on the results of previous steps, thus improving the system's responsiveness and overall task performance.

\section{Details of the Evaluation Benchmarks}
\label{app:benchmark-details}

\paragraph{MultiWOZ 2.4.} MultiWOZ 2.4 \cite{ye-etal-2022-multiwoz} is a multi-domain TOD dataset designed to evaluate dialogue systems' ability to handle complex conversations across multiple domains such as hotel booking, restaurant reservations, and transportation. We employ two different metrics during our TOD evaluations MultiWOZ: \textit{Success Rate}, which assesses whether all user-requested information related to the entity is successfully provided and Joint Goal Accuracy (JGA) which measures the accuracy of predicted dialogue states, reflecting the system's ability to track user intents. During our zero-shot evaluations, we used its test set that contains 999 samples and incorporated AutoTOD prompt \cite{xu-etal-2024-rethinking} with slight modifications, thereby generating system responses analogous to those produced in a chat-based inference setting.

\paragraph{API-Bank.} API-Bank \cite{li-etal-2023-apibank} is designed to evaluate tool-augmented LAs, focusing on their ability to plan, retrieve, and invoke APIs effectively. It includes 314 tool-use dialogues and 753 API calls, with two evaluation levels: Level 1 (L-1), which tests the accuracy of invoking a known API based on a given query, and Level 2 (L-2), which assesses the retrieval and invocation of APIs from a candidate list, simulating real-world scenarios with multiple API options. By addressing these challenges, API-Bank advances the understanding and enhancement of tool-augmented reasoning in LLMs. During evaluations, we used the official evaluation code from the repository of previous works \cite{Lin2024Hammer}.

\paragraph{Berkeley Function Calling Leaderboard.} In addition to API-Bank, we also used BFCL V3\footnote{\url{https://gorilla.cs.berkeley.edu/blogs/13_bfcl_v3_multi_turn.html}} \cite{patil2023gorilla} which provides a diverse evaluation framework for assessing the models' ability to perform function calls across various objectives. It includes more than 1,800 test cases that span tasks such as simple functions, multiple functions, and parallel functions for Python and other environments such as REST APIs and JavaScript. Models are evaluated using two primary metrics: (i) Abstract Syntax Tree (AST) accuracy, which ensures syntactic correctness by verifying function structures, parameters, and types against predefined documentation and (ii) Executable Function Accuracy, which evaluates whether generated functions execute correctly and produce the expected outputs, emphasizing real-world applicability. In our experiments, we employed the official repository released by authors and followed the provided instructions to get model results.

\section{Baseline Model Overviews Used in Experiments}
\label{app: model-overview} 
In this section, we provide an overview of the models used in our experiments, including their brief descriptions, checkpoints, and the training re-production code references.

\subsection{Base Models}

\paragraph{Llama 3.1.} The Llama (Large Language Model Meta AI) \cite{Dubey2024TheL3-llama3} family is a set of open-source language models from Meta AI, ranging from 7 to 405 billion parameters. It trained on a large corpus of web content, academic texts, and books, they excel at reasoning, question-answering, and code generation. Their architecture supports efficient fine-tuning and deployment. In our experiments, we use Llama-3.1-8B-Instruct\footnote{\url{https://huggingface.co/meta-llama/Llama-3.1-8B-Instruct}}, released in July 2024, which offers improved multilingual capabilities, longer context windows, and state-of-the-art performance in general knowledge, math, and tool usage

\paragraph{Mistral v03.} Mistral 7B \cite{jiang2023mistral} is one of the state-of-the-art, open-source LLMs produced by Mistral AI. It employs innovative mechanisms such as grouped-query and sliding window attention, which enable efficient processing of longer sequences and faster inference times. In our experiments, we use Mistral-7B-Instruct-v0.3\footnote{\url{https://huggingface.co/mistralai/Mistral-7B-Instruct-v0.3}}, released on May 22, 2024, and available on Hugging Face.

\subsection{TOD Models}
 
\paragraph{LDST.} LDST (LLM-driven Dialogue State Tracking) \cite{feng2023towards} is an approach that overcomes the limitations of proprietary models in state tracking by leveraging a fine-tuned LLaMa 7B model. The approach combines a novel assembled domain-slot instruction tuning technique with parameter-efficient strategies, enabling resource-efficient performance that tries matches larger models. During our experiments and to fine-tune LDST we used the provided checkpoints and implementation details for LDST are available in their public repository \footnote{\url{https://github.com/WoodScene/LDST}}.

\paragraph{Fnc-TOD.} FNC-TOD (Function-Calling for Task-Oriented Dialogue) focuses on DST in LLMs through function calling mechanisms. The method conceptualizes domain schemas as functions and embeds function specifications within the system prompt. This approach achieved improved conversation state tracking in task-oriented dialogues using a fine-tuned Llama-2-13b-chat-hf model, trained on a focused dataset of 7,200 task-oriented dialogues spanning 36 domains. For our experiments, we utilized the authors' publicly released Zekunli/FncTOD-Llama-13b model available on Huggingface \footnote{\url{https://huggingface.co/Zekunli/FncTOD-Llama-13b}}.

\paragraph{NC-Latent-TOD.} This work introduces an unsupervised approach to TOD systems that operates solely with API schemas and unlabeled dialogues, eliminating the need for costly turn-level annotations. The system generates pseudo-labels for API calls and system actions while using a Hard-Expectation maximization approach with LLM predictions for iterative fine-tuning, enhanced by a noisy-channel reranking method \cite{king-flanigan-2024-unsupervised}. During our experiments, we used two different models nc-latent-tod-step-2-final\footnote{\url{https://huggingface.co/Brendan/nc-latent-tod-step-2-final}} and tod-zero-bqag3oyb-32000\footnote{\url{https://huggingface.co/Brendan/tod-zero-bqag3oyb-32000}} shared by the authors.

\subsection{Language Agents}

\paragraph{CodeAct-Agent.} CodeAct \cite{wang2024executable} is a framework that enables LLM agents to generate and execute Python code as actions to interact with environment, rather than being limited to JSON or structured text formats. By integrating a Python interpreter, it allows agents to dynamically adjust their actions based on execution results, leverage existing Python packages, and utilize programming constructs like loops and conditionals for complex operations. authors developed CodeActAgent by fine-tuning both Mistral 7B and Llama2 7B models on the CodeAct-Instruct dataset. For our experiments, we utilized the authors' officially released CodeActAgent-Mistral-7b-v0.1 model, available on Huggingface \footnote{\url{https://huggingface.co/xingyaoww/CodeActAgent-Mistral-7b-v0.1}}.

\paragraph{Granite-20B.} This work introduces Granite-20B, an open-source LLM, specifically designed for function calling capabilities. The model is trained using a multi-task approach on seven core function calling tasks: Nested Function Calling, Function Chaining, Parallel Functions, Function Name Detection, Parameter-Value Pair Detection, Next-Best Function, and Response Generation. We used the offical model weights granite-20b-code-instruct-8k provided in Huggingface\footnote{\url{https://huggingface.co/ibm-granite/granite-20b-code-instruct-8k}}.

\paragraph{Hammer2.0-7B.} Hammer \cite{Lin2024Hammer} is a small scale model family up to 7B parameter models designed for on-device function calling and addresses generalization challenges in function calling through two key innovations: an irrelevance-augmented dataset that enhances models' ability to identify inappropriate functions, and a function masking technique that reduces naming-based misinterpretations by focusing on function descriptions. Built by fine-tuning the xLAM-function-calling dataset\footnote{\url{https://huggingface.co/datasets/Salesforce/xlam-function-calling-60k}} with 7,500 additional instances for irrelevance detection, Hammer achieves state-of-the-art performance on BFCL Benchmark. For our experiments, we utilized the official Hammer 2.0 model weights available on Huggingface\footnote{\url{https://huggingface.co/MadeAgents/Hammer2.0-7b}}, along with training it from scratch for reproducibility using provided public repository and training scripts\footnote{\url{https://github.com/MadeAgents/Hammer}}.

\paragraph{ToolAce 8B.} This work introduces ToolACE \cite{Liu2024ToolACE}, an automated pipeline for generating high-quality function-calling training data. The system features a self-evolution synthesis process that curates a pool of 26,507 diverse APIs, coupled with a multi-agent dialogue generation system and a dual-layer verification process for ensuring data accuracy. Using data generated and fine-tuning on Llama-3.1-8B-Instruct, ToolACE achieve top results on the BFCL Leaderboard. We used the official Huggingface checkpoint\footnote{\url{https://huggingface.co/Team-ACE/ToolACE-8B}} and dataset\footnote{\url{https://huggingface.co/datasets/Team-ACE/ToolACE}}.

\section{Human Validation for Generated CRA Dataset}
\label{human-validation}
To analyze the quality of generated conversations, we implemented a systematic random sampling approach. From the generated dataset, we randomly selected 100 dialogue instances for validation. We conducted the evaluation against a predefined set of 51 available functions, covering transportation, booking, entertainment, and utility services. We scrutinized each function’s schema, including its parameters and expected usage, to ensure compliance. We asked a senior Computer Science student to evaluate these generated samples across four key dimensions:
\begin{itemize}
    \item \textbf{Undefined Function Call:} Validating API names and parameters against the predefined function list to identify undefined functions or invalid arguments.
    \item \textbf{Incorrect Argument Type:} Checking argument structures to ensure compliance with the function schemas.
    \item \textbf{Argument Hallucination:} Detecting unnecessary or irrelevant arguments misaligned with the conversation context.
    \item \textbf{Low-Quality Reasoning and Planning:} Identifying logical gaps in though steps or unnecessary API calls in ReAct structure.
\end{itemize}
We asked for a binary score (1 for no errors, 0 for detected issues) for each generated dialogue and provided mandatory feedback for any errors. Our evaluation of 100 dialogues showed a 9\% error rate, mostly in restaurant reservations where key details like the restaurant name or dining time were missing. These errors stemmed from Argument Hallucination and Low-Quality Reasoning. Results, including dialogue IDs, scores, and feedback, were systematically collected to identify areas for improvement.






\begin{figure*}[t!]
\includegraphics[width=\textwidth]{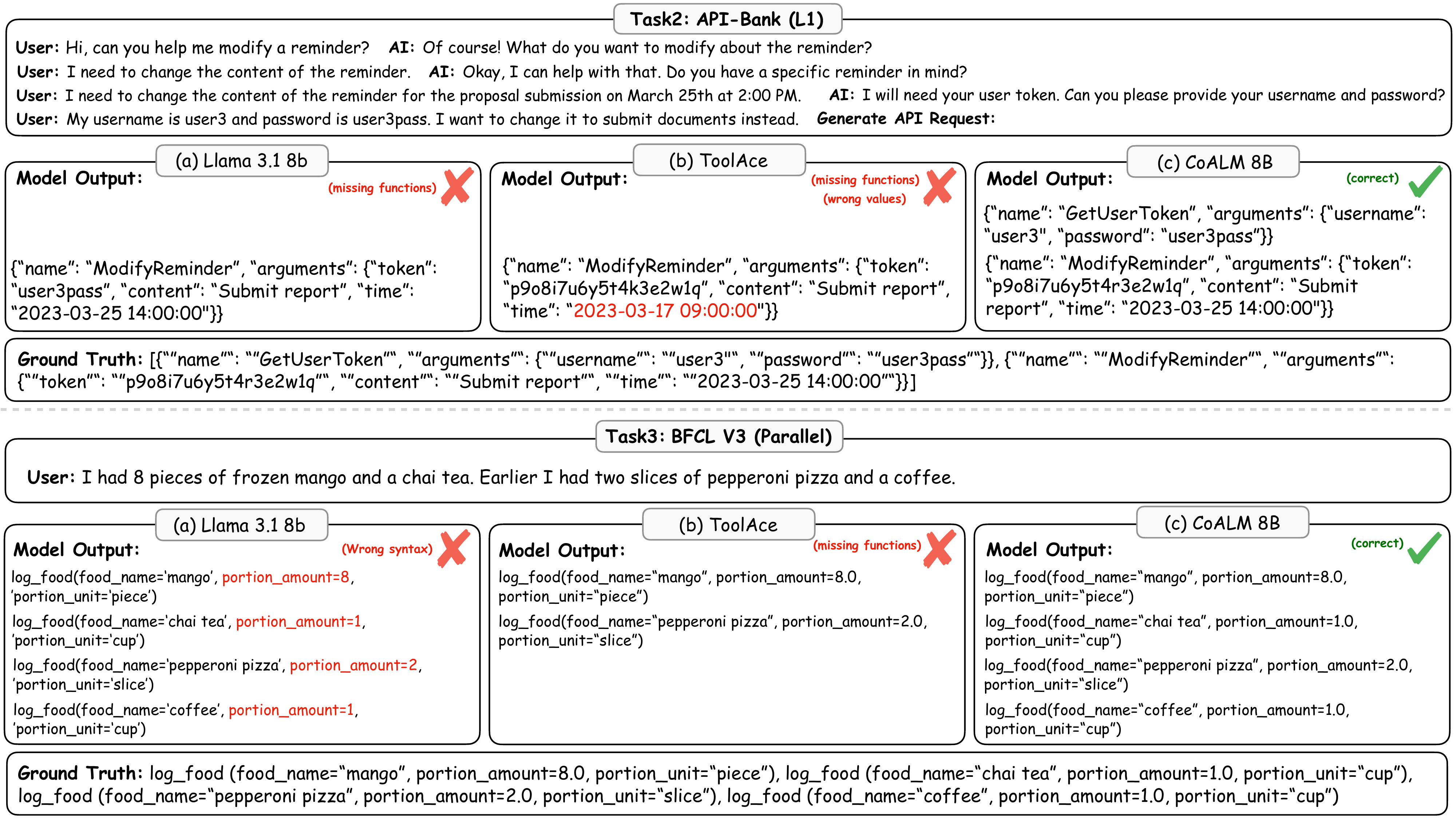}
\caption{\textbf{Error Analysis of Function-Calling Results.} Illustrated performance comparison on function calling benchmarks API-Bank L1 (top) and BFCL V3 parallel function call (bottom). Results demonstrate CoALM's consistent performance compared to other baselines.}
\label{fig:error-analysis}
\end{figure*}
\clearpage

\begin{figure*}[!ht]
\begin{tcolorbox}[colback=gray!5!white,colframe=black!95!black,title=\textbf{\small{SNIPS SFT Sample} | \textbf{Format:} Dialogue State Tracking}] 
\small \textcolor{darkgreen}{\textbf{Instruction:}} \\
You are a helpful assistant who is assigned to find the intents shown by the user on 7 domains - GetWeather, AddToPlaylist, SearchScreeningEvent, BookRestaurant, SearchCreativeWork, RateBook, PlayMusic. \\
 \\
The user can seek for BookRestaurant by slots - poi, restaurant\_type, served\_dish, timeRange, party\_size\_number, restaurant\_name, state, country, party\_size\_description, sort, city, spatial\_relation, cuisine, facility. \\
The user can seek for GetWeather by slots - condition\_temperature, geographic\_poi, current\_location, timeRange, condition\_description, state, country, city, spatial\_relation. \\
The user can seek for SearchCreativeWork by slots - object\_type, object\_name. \\
The user can seek for PlayMusic by slots - track, playlist, service, genre, year, album, music\_item, sort, artist. \\
The user can seek for SearchScreeningEvent by slots - movie\_name, location\_name, timeRange, object\_type, movie\_type, object\_location\_type, spatial\_relation. \\
The user can seek for RateBook by slots - rating\_value, rating\_unit, object\_type, object\_select, object\_part\_of\_series\_type, best\_rating, object\_name.
Do not capture any other slots! \\
 \\
\# Task \\
You will be provided with an user utterance. You must find all the user intents and output them in JSON format. \\
 \\
\# Sample Output \\
{"domain": "AddToPlaylist", "slot\_values": {"music\_item": "abc", "artist": "xyz"}} \\
 \\
\small \textcolor{darkgreen}{\textbf{Input:}} \\
User: Book a table at a restaurant in Portugal with parking for me and bonnie in 19 minutes\\
 \\
\small \textcolor{mypurple}{\textbf{Output:}} \\
System: {"domain": "BookRestaurant", "slot\_values": {"restaurant\_type": "restaurant", "country": "Portugal", "facility": "parking", "party\_size\_description": "me and bonnie", "timeRange": "in 19 minutes"}}"
\end{tcolorbox}

\vspace{-0.25cm}
\caption{SNIPS fine-tuning sample example.}
\label{tab:snips-dst}
\end{figure*}

\newpage
\begin{figure*}[!ht]
\begin{tcolorbox}[colback=gray!5!white,colframe=black!95!black,title=\textbf{\small{Hammer SFT Sample} | \textbf{Format:} Function Calling}] 
\small \textcolor{darkgreen}{\textbf{Instruction:}} \\
{[BEGIN OF TASK INSTRUCTION]} \\
You are a tool calling assistant. In order to complete the user's request, you need to select one or more appropriate tools from the following tools and fill in the correct values for the tool parameters. Your specific tasks are: \\
1. Make one or more function/tool calls to meet the request based on the question. \\
2. If none of the function can be used, point it out and refuse to answer. \\
3. If the given question lacks the parameters required by the function, also point it out. \\
{[END OF TASK INSTRUCTION]} \\
 \\
{[BEGIN OF AVAILABLE TOOLS]} \\
{[{"name": "LxOm64zLyg", "description": "Gets hourly weather forecast information for given geographical coordinates using the RapidAPI service.", "parameters": {"TDpjPd": {"description": "The latitude of the geographical location.", "type": "int", "default": 46.95828}, "78th2U3lFj": {"description": "The longitude of the geographical location.", "type": "int", "default": 10.87152}}}, {"name": "WoDdNSe7e7K5", "description": "Fetches weather updates for a given city using the RapidAPI Weather API.", "parameters": {"LzZsvxUC": {"description": "The name of the city for which to retrieve weather information.", "type": "str", "default": "London"}}}, {"name": "CBrCNmwOERb", "description": "Fetches the hourly weather forecast for a given location using the RapidAPI service.", "parameters": {"TDEJ.ZwMt": {"description": "The name of the location for which to retrieve the hourly weather forecast.", "type": "str", "default": "Berlin"}}}, {"name": "1YTQVXkwLY", "description": "Returns an air quality forecast for a given location.", "parameters": {"2bkgDA": {"description": "The latitude of the location for which the air quality forecast is to be retrieved.", "type": "int", "default": "35.779"}, "DQi.ReZ16": {"description": "The longitude of the location for which the air quality forecast is to be retrieved.", "type": "int", "default": "-78.638"}, "hF.1": {"description": "The number of hours for which the forecast is to be retrieved (default is 72).", "type": "int", "default": "72"}}}]} \\
{[END OF AVAILABLE TOOLS]} \\
 \\
{[BEGIN OF FORMAT INSTRUCTION]} \\
The output MUST strictly adhere to the following JSON format, and NO other text MUST be included. \\
The example format is as follows. Please make sure the parameter type is correct. If no function call is needed, please directly output an empty list '{[]}' \\
{[ \\
{"name": "func\_name1", "arguments": {"argument1": "value1", "argument2": "value2"}}, \\
... (more tool calls as required) \\
]} \\
{[END OF FORMAT INSTRUCTION]} \\
\\
\small \textcolor{darkgreen}{\textbf{Input:}} \\
{[BEGIN OF QUERY]} \\
What are the current weather conditions in Sydney? \\
{[END OF QUERY]} \\
 \\
\small \textcolor{mypurple}{\textbf{Output:}} \\
{[{"name": "WoDdNSe7e7K5", "arguments": {"LzZsvxUC": "Sydney"}}]}
\end{tcolorbox}

\vspace{-0.25cm}
\caption{Hammer fine-tuning sample example.}
\label{tab:hammer-api-2}
\end{figure*}

\newpage
\begin{figure*}[!ht]
\begin{tcolorbox}[colback=gray!5!white,colframe=black!95!black,title=\textbf{\small{SGD Instruction Sample} | \textbf{Format:} Action Optimization}] 
\small \textcolor{darkgreen}{\textbf{Instruction:}} \\
{[BEGIN OF TASK INSTRUCTION]}\\
You are a helpful conversational assistant who can perform API function calling. \\
Your goal is to understand user queries and respond using the appropriate API functions. \\
In order to complete the user's request, you need to select a tool from the following functions and fill in the correct values for the function parameters. \\
Your specific tasks are: \\
1. Analyze the user’s query within the given dialogue context to identify their intent and relevant details. \\
2. Make a function/tool call and provide the necessary arguments to meet the request based on the user query. \\
3. Formulate a natural and coherent response, guiding the conversation towards resolving the user’s request. \\
{[END OF TASK INSTRUCTION]} \\
 \\
{[BEGIN OF AVAILABLE TOOLS]} \\
1. FindBus(from\_location, to\_location, leaving\_date) \\
2. ReserveCar(pickup\_location, pickup\_date, pickup\_time, dropoff\_date, type) \\
. \\
. \\
. \\
50. FindProvider(city, type) \\
51. GetRide(destination, number\_of\_riders, shared\_ride) \\
{[END OF AVAILABLE TOOLS]} \\
 \\
{[BEGIN OF FORMAT INSTRUCTION]} \\
The output MUST strictly adhere to the following structured text format. \\
Example Output API Call Format: \\
function\_name(argument1=value1, argument2=value2, ...) \\
{[END OF FORMAT INSTRUCTION]} \\
 \\
{[BEGIN OF CONVERSATION HISTORY]} \\
User: I'm looking for events happening in New York this weekend. \\
Thought: The user wants to find events in a specific location and timeframe.  \\
I'll need to call the FindEvents API with the appropriate category and city. \\
Action: FindEvents(category='all', city\_of\_event='New York') \\
Observation: {'events': [{'name': 'Jazz Festival', 'date': '2023-10-07', 'location': 'Central Park'}, {'name': 'Food Truck Rally', 'date': '2023-10-08', 'location': 'Union Square'}]} \\
Thought: I have the event details now. I should summarize the events for the user. \\
System: This weekend in New York, you can attend the Jazz Festival at Central Park on October 7th and the Food Truck Rally at Union Square on October 8th. Would you like more information about any of these events? \\
User: Yes, can you tell me more about the Jazz Festival? \\
Thought: I need to provide more details about the Jazz Festival. I don't need an API call for that. \\
System: The Jazz Festival in Central Park will feature various artists performing live jazz music throughout the day. It starts at 11 AM and goes until 8 PM. Would you like to know how to get tickets? \\
{[END OF CONVERSATION HISTORY]} \\
 \\
\small \textcolor{darkgreen}{\textbf{Input:}} \\
User: Yes, please tell me how to get tickets for the Jazz Festival. \\
Thought: The user is interested in purchasing tickets. I need to call the BuyEventTickets API for the Jazz Festival. \\
 \\
\small \textcolor{mypurple}{\textbf{Output:}} \\
Action: BuyEventTickets(event\_name='Jazz Festival', number\_of\_seats=2, date='2023-10-07', city\_of\_event='New York')\\
\end{tcolorbox}

\vspace{-0.25cm}
\caption{SGD fine-tuning sample example targeting function optimization.}
\label{tab:sgd-sft-action}
\end{figure*}

\newpage
\begin{figure*}[!ht]
\begin{tcolorbox}[colback=gray!5!white,colframe=black!95!black,title=\textbf{\small{SGD Instruction Sample} | \textbf{Format:} System Response Optimization}] 
\small \textcolor{darkgreen}{\textbf{Instruction:}} \\
{[BEGIN OF TASK INSTRUCTION]}\\
You are a helpful conversational assistant specializing in understanding user queries and providing accurate, reasoned responses. \\
Your goal is to analyze the user's input, reason about their intent and needs, and provide a coherent and contextually appropriate system response. \\
Your specific tasks are: \\
- Ensure your response is informative and contextually relevant, guiding the conversation toward successful task completion. \\
- Analyze the user's input in the context of the conversation history (if available) to identify their intent and relevant details. \\
- Use logical reasoning to determine the most suitable response, considering the user's needs and the dialogue context. \\
- Generate a natural and coherent system response to address the user’s request or query effectively. \\
{[END OF TASK INSTRUCTION]} \\
\\
{[BEGIN OF FORMAT INSTRUCTION]} \\
The output MUST strictly adhere to the following structured text format. \\
Example Output API Call Format: \\
function\_name(argument1=value1, argument2=value2, ...) \\
{[END OF FORMAT INSTRUCTION]} \\
 \\
{[BEGIN OF CONVERSATION HISTORY]} \\
User: I'm looking for events happening in New York this weekend. \\
Thought: The user wants to find events in a specific location and timeframe.  \\
I'll need to call the FindEvents API with the appropriate category and city. \\
Action: FindEvents(category='all', city\_of\_event='New York') \\
Observation: {'events': [{'name': 'Jazz Festival', 'date': '2023-10-07', 'location': 'Central Park'}, {'name': 'Food Truck Rally', 'date': '2023-10-08', 'location': 'Union Square'}]} \\
Thought: I have the event details now. I should summarize the events for the user. \\
System: This weekend in New York, you can attend the Jazz Festival at Central Park on October 7th and the Food Truck Rally at Union Square on October 8th. Would you like more information about any of these events? \\
User: Yes, can you tell me more about the Jazz Festival? \\
Thought: I need to provide more details about the Jazz Festival. I don't need an API call for that. \\
System: The Jazz Festival in Central Park will feature various artists performing live jazz music throughout the day. It starts at 11 AM and goes until 8 PM. Would you like to know how to get tickets? \\
{[END OF CONVERSATION HISTORY]} \\
 \\
\small \textcolor{darkgreen}{\textbf{Input:}} \\
User: Yes, please tell me how to get tickets for the Jazz Festival. \\
Thought: The user is interested in purchasing tickets. I need to call the BuyEventTickets API for the Jazz Festival. \\
Action: BuyEventTickets(event\_name='Jazz Festival', number\_of\_seats=2, date='2023-10-07', city\_of\_event='New York') \\
Observation: {'status': 'success', 'message': 'You have successfully purchased 2 tickets for the Jazz Festival.'} \\
Thought: The tickets were successfully purchased. I need to confirm this with the user. \\
 \\
\small \textcolor{mypurple}{\textbf{Output:}} \\
System: You have successfully purchased 2 tickets for the Jazz Festival in Central Park! Enjoy the music! \\
\end{tcolorbox}

\vspace{-0.25cm}
\caption{SGD fine-tuning sample example targeting system response optimization.}
\label{tab:sgd-sft-response}
\end{figure*}

\newpage
\begin{table*}[!t]
\scriptsize
\centering
\ttfamily
\begin{tabular}{p\linewidth}
\toprule
You are an advanced AI assistant specializing in conversational dialogues. \\
You have access to a variety of services and APIs to assist users with their requests and your goal is to provide helpful and informative responses to user queries and commands. \\
You can interact with multiple services and APIs to fulfill user requests. \\
Your responses should be natural, informative, and tailored to the user's needs. \\
 \\
\# Task Information: \\
- You are asked to create a dataset in the format: User - Thought1 - API - API Input Arguments - API Result - Thought2 - System, or User - Thought - System.  \\
- For the given \# User Input, generate a multi-turn dialogue that follows this format, with each turn exhibiting realistic context reasoning, thought processes, and API interaction where applicable.  \\
- The dialogues should be converted to follow a specific \# Output Format, which includes reasoning on whether an API call is needed or if the system can respond directly. \\
- If the system decides that an API call is necessary, use this format: User - Thought1 - API - API Input Arguments - API Result - Thought2 - System. \\
- Call the right API from \# Avaliable Functions and provide the necessary input arguments to fulfill the user's request. \\
- If you think a function argument is not necessary, you can skip it. Don't provide unnecessary arguments and None values. \\
- Ensure that the API calls are used logically and that the dialogue remains coherent and natural throughout the exchange.  \\
- If the system determines that an API call is not necessary, use this format: User - Thought - System. \\
- Include intermediate thoughts where appropriate to capture the model's internal reasoning, and clearly separate the different components of the format. \\
\\
\# Available Functions: \\
1. CheckBalance(account\_type) \\
2. TransferMoney(account\_type, amount, recipient\_account\_name) \\
. \\
. \\
. \\
52. FindAttractions(location) \\
53. GetWeather(city) \\
 \\
\# Output Format: \\
- If an API Call is Needed: \\
\quad	    User: [User Input] \\
\quad	    Thought1: [I need to call an API] \\
\quad	    API Name: [API Call Name: CheckBalance(), TransferMoney(),...,  FindAttractions(), GetWeather()] \\
\quad	    API Input: [The input parameters for the API] \\
\quad	    API Result: [API output result] \\
\quad	    Thought2: [2nd thought after API Result that ensure if the information is enough before the system response] \\
\quad	    System: [Your system response here] \\
 \\
- If an API Call is Not Needed: \\
\quad	    User: [User Input] \\
\quad	    Thought: [I don't need an API and I want to respond to the user] \\
\quad	    System: [Your system response here] \\
 \\
Below is an example of the format: \\
 \\
\# Example: \\
User: Add Don and Sherri to my "Meditate to Sounds of Nature" playlist.   \\
Thought1: The user wants to add two artists to a specific playlist. I'll need to extract the playlist name and the artists' names.  \\ 
API Name: AddToPlaylist   \\
API Input: {'playlist\_name': 'Meditate to Sounds of Nature', 'artists': Don Sherri}   \\
API Result: {'status': 'success', 'message': 'Don and Sherri have been added to your playlist.'}   \\
Thought2: The API call was successful, and the artists were added to the playlist. I need to inform the user about this.   \\
System: Don and Sherri have been successfully added to your "Meditate to Sounds of Nature" playlist. Enjoy your music!   \\
 \\
User: Thank you for adding them. I think I can add more songs to the playlist. \\
Thought: I don't need an API call for that, I want to respond to the user.  \\
System: Would you like to add more songs?   \\
 \\
User: Yes, please add "Calm River" to the playlist.   \\
Thought1: The user wants to add another song to the playlist. I'll note the song title and proceed with the request.   \\
API Name: AddToPlaylist   \\
API Input: {'playlist\_name': 'Meditate to Sounds of Nature', 'songs': Calm River}   \\
API Result: {'status': 'success', 'message': 'Calm River has been added to your playlist.'}   \\
Thought2: The song was successfully added. I should let the user know.   \\
System: "Calm River" has been successfully added to your "Meditate to Sounds of Nature" playlist. Would you like to do anything else?   \\
 \\
User: No, that's all for now. Thank you!   \\
Thought: The user is satisfied and doesn't need further assistance. I should acknowledge their gratitude.   \\
System: You're welcome! If you need anything else, feel free to ask. Enjoy your relaxing music! \\
 \\
\# User Input:\\
\bottomrule
\end{tabular}
\caption{API-driven conversation prompt to generate SGD fine-tuning samples with GPT4-o. Only functions used across all dialogues are included due to their large number.}
\label{tab: prompt}
\end{table*}

\end{document}